\documentclass[10pt,twocolumn,letterpaper]{article}

\usepackage{iccv}              

\newcommand{\IT}[1]{``{\fontfamily{cmtt}\selectfont #1}''}
\usepackage[T1]{fontenc}
\usepackage[htt]{hyphenat}
\hyphenchar\font=`\-

\usepackage{array}
\newcommand{\figsref}[1]{Figs.~\ref{#1}}

\definecolor{iccvblue}{rgb}{0.21,0.49,0.74}
\usepackage[pagebackref,breaklinks,colorlinks,allcolors=iccvblue]{hyperref}

\def\paperID{3} 
\def\confName{ICCV}
\def\confYear{2025}

\title{Embedding Font Impression Word Tags Based on Co-occurrence}

\author{
Yugo Kubota \quad Seiichi Uchida \\
Kyushu University, Japan \\
\tt\small{\{yugo.kubota, uchida\}@human.ait.kyushu-u.ac.jp}
}

\begin{document}
\maketitle

\begin{abstract}
Different font styles (i.e., font shapes) convey distinct impressions, indicating a close relationship between font shapes and word tags describing those impressions. 
This paper proposes a novel embedding method for impression tags that leverages these shape–impression relationships. 
For instance, our method assigns similar vectors to impression tags that frequently co-occur in order to represent impressions of fonts, whereas standard word embedding methods (e.g., BERT and CLIP) yield very different vectors. 
This property is particularly useful for impression-based font generation and font retrieval. Technically, we construct a graph whose nodes represent impression tags and whose edges encode co-occurrence relationships. 
Then, we apply spectral embedding to obtain the impression vectors for each tag. 
We compare our method with BERT and CLIP in qualitative and quantitative evaluations, demonstrating that our approach performs better in impression-guided font generation.
\end{abstract}

\section{Introduction}
\label{sec:intro}
\begin{figure}[t] 
    \centering
    \includegraphics[width=\linewidth]{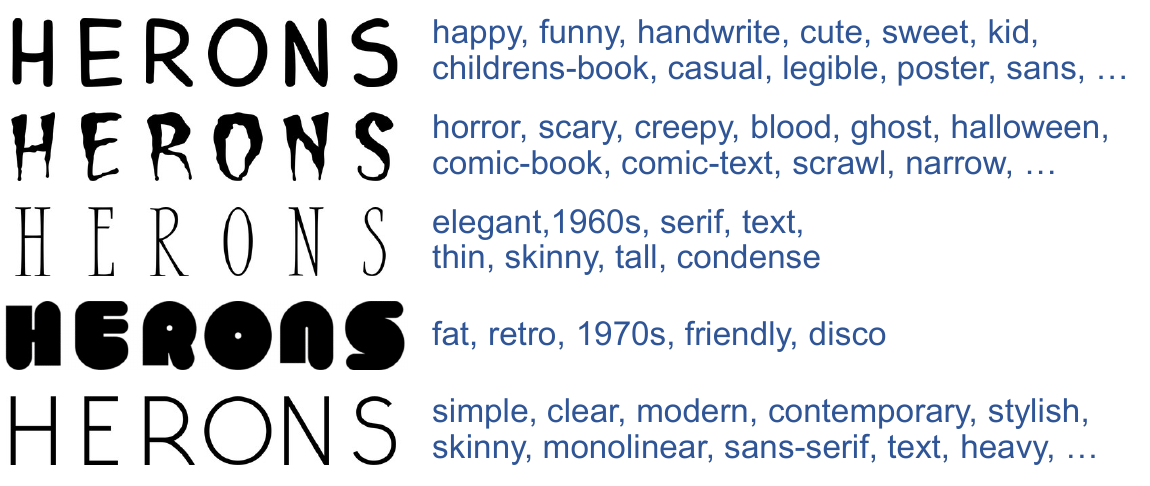}
    \caption{Examples of fonts and their impression word tags. The six characters ``HERONS'' are shown as they represent diverse stroke variations within each font.}
    \label{fig:font_ex}
\end{figure}

Different fonts convey different impressions. 
\cref{fig:font_ex} shows several examples of fonts and their associated impression word tags (hereafter referred to as \emph{impression tags}) from the MyFonts dataset~\cite{MyFonts}. 
These impression tags are assigned to each font by users via crowdsourcing.
The impressions conveyed by a font can vary significantly depending on its shape.
For example, the font in the top row conveys positive and casual impressions and may be suitable for printing the word ``HAPPY.'' 
In contrast, the font in the second row conveys negative impressions and may be unsuitable for the same word ``HAPPY.''
\par

Understanding the relationship between font shapes and impressions is crucial from both theoretical and practical perspectives. 
From a theoretical standpoint, uncovering this relationship deepens our understanding of visual design principles and shape perception.
From a practical perspective, understanding this relationship has wide-ranging applications in areas such as branding, design evaluation, computer-aided font generation, and other typographic tasks.
\par

\begin{figure}[t] 
    \centering
    \includegraphics[width=\linewidth]{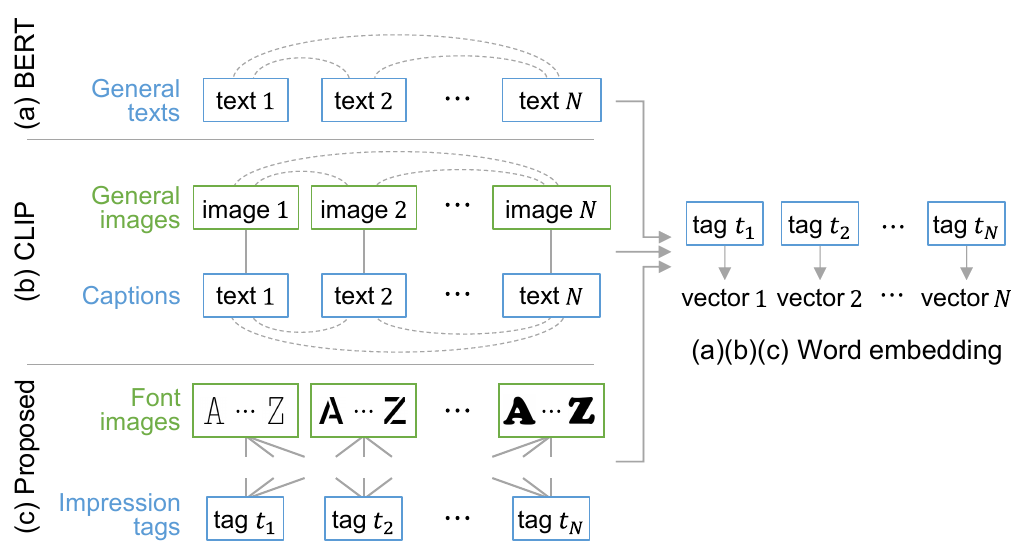}
    \caption{Conceptual comparison of word embedding methods, where (c) shows the proposed method.}
    \label{fig:scheme}
\end{figure} 

To understand this relationship, a key challenge lies in the numerical representation (\ie, vector representation) of font impressions. 
Traditional approaches typically rely on word embedding methods such as word2vec~\cite{word2vec}, with more recent work introducing methods like BERT~\cite{BERT} and CLIP~\cite{CLIP}.
However, as shown in \figsref{fig:scheme}(a) and (b), these existing methods treat impression tags as ordinary words in general texts, rather than as impressions grounded in font shapes.
Indeed, as shown in \cref{fig:font_ex}, these impressions are inherently tied to the visual characteristics of fonts. 
For example, in general usage, the tags \IT{elegant} and \IT{thin} differ in meaning, yet both may be assigned to the same font---as seen in the third example---due to similarities from a visual perspective.
In such cases, their embedding vectors should be similar and reflect their shared grounding in that particular font shape.
\par

\begin{figure}[t] 
    \centering
    \includegraphics[width=\linewidth]{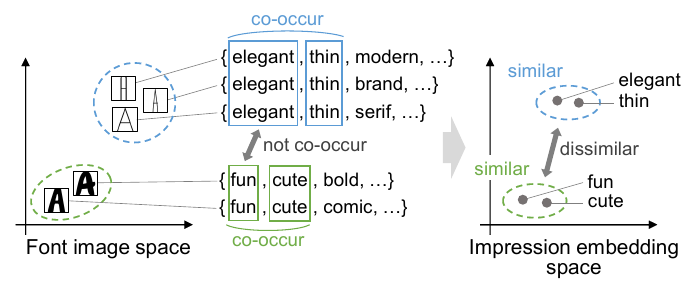}
    \caption{
    Impression tags \IT{elegant} and \IT{thin} frequently co-occur and are generally associated with fonts with similar shapes. Our co-occurrence-based method assigns similar ``impression embedding vectors'' to them. Consequently, their similarity reflects the visual similarity of fonts.}
    \label{fig:illustration}
\end{figure}

In this paper, we propose a co-occurrence-based embedding method for impression tags. 
As shown in \cref{fig:scheme}(c), our method addresses the limitations of conventional word embedding methods by focusing on how impressions arise from font shapes.
Since impressions are conveyed by font shapes, we assume that impression tags frequently associated with the same fonts (\ie, co-occurring tags) should be represented by similar embedding vectors.
For example, as illustrated in \cref{fig:illustration}, two tags \IT{elegant} and \IT{thin} tend to co-occur in fonts with similar shapes. Consequently, by focusing on tag co-occurrence and assigning similar vectors to frequently co-occurring tags, we can obtain vector representations of impression tags that reflect the visual characteristics of fonts.
\par

The proposed embedding method has several notable features. 
First, our embedding vectors are tightly linked to font shapes rather than general texts and general images, and therefore behave very differently from the vectors produced by BERT and CLIP. 
Second, our vectors are particularly suitable for modeling how font shapes convey specific impressions; this feature is useful when regulating neural network-based font generation processes by specifying the desired impressions.
\par

We evaluate the proposed co-occurrence-based embedding method for impression tags through four experiments:
\begin{itemize}
    \item First, we visualize the distribution of impression tags in the embedding space and observe their clusters.
    \item Second, we examine the correspondence between impression tag embeddings and actual font images, and find that tags that are similar in the proposed embedding space tend to be associated with similar fonts.
    \item Third, we demonstrate how the impression tag--font correspondence can help identify ``missing tags'' (\ie, tags that should be assigned but are absent) and ``noisy tags'' (\ie, tags that are incorrectly assigned).    
    \item Finally, we conduct a conditional font generation experiment, in which the embedding vectors guide the generation of fonts that reflect specified impressions.
\end{itemize}

\section{Related Work}
\subsection{Font Impression Analysis}
Researchers have long studied how font shapes affect the subjective impressions people form.
Early work dates back to the 1920s~\cite{davis1933determinants, franken1923study}, and numerous subsequent studies have demonstrated consistent associations between specific fonts and their impressions~\cite{Morrison2006Educational, Shaikh2006Perception, Ying2010Typeface}.
In practical applications, these associations serve as effective tools for conveying nonverbal cues, such as brand and product identities in marketing~\cite{henderson2004impression, VELASCO2014, Maísa2004typefaces}, or emotional tone in text-based communication~\cite{TypefaceEmotion, Emotype, CHUJO20242Exploring, EmoScribe}.
Traditionally, much of this work has been based on small-scale user studies focusing on relatively limited font sets and impression vocabularies.
\par

Recently, studies on font–impression relationships have increasingly focused on large-scale datasets and machine-learning-based approaches~\cite{o2014exploratory, MyFonts}.
Several studies have leveraged these large-scale datasets alongside multimodal models to learn font-impression correspondences~\cite{kulahcioglu2020fonts, choi2019assist, CrossAE}.
In particular, following the success of CLIP~\cite{CLIP}, contrastive learning has been used to embed fonts and impressions into a shared space that captures their correspondence~\cite{ImpressionCLIP, FontCLIP}.
While previous methods vectorize both font images and impressions, our method emphasizes vectorizing impressions based on tag co-occurrence within fonts. 
\par

\begin{figure*}[t] 
    \centering      
    \includegraphics[width=\linewidth]{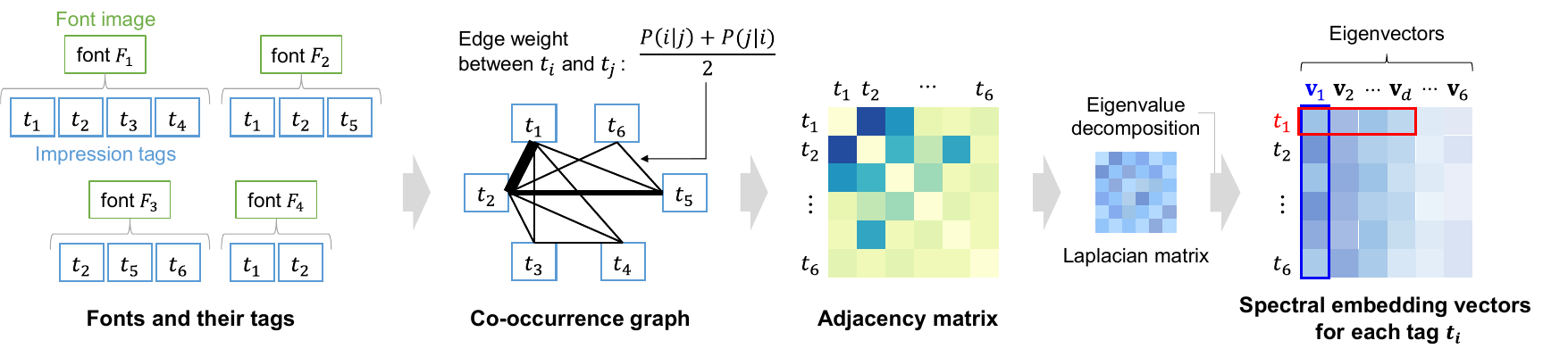}
    \caption{Overview of our impression tag embedding method. For simplicity,  the tag vocabulary size is set to $N = 6$ in this illustration.}
    \label{fig:overview}
\end{figure*} 

\subsection{Generating Fonts with Impressions}
\label{sec:review-gen}
From a practical perspective, the relationship between font shapes and impressions is important for impression-guided font generation. In general, machine learning-based font generation follows a few-shot paradigm that requires a couple of pre-designed characters as style references~\cite{Zhang2018Separating, Gharbi2024VecFusion, Fu2024Generate, Xie2021DG-Font, Yang2024FontDiffuser}. In contrast, impression-guided font generation seeks to generate fonts by specifying only the expected impressions and thus can eliminate the need for pre-designed characters.
This allows non-expert users to intuitively generate new fonts by specifying impressions that describe their desired font style, thereby making the process user-friendly.

Vectorizing impressions is a key step for providing style-related information to impression-guided generative models.
Matsuda \etal~\cite{matsuda2021impressions2font, matsuda2022font} introduced generative adversarial approaches that utilize word2vec~\cite{word2vec} for vectorizing impression tags.
Furthermore, some approaches employ CLIP~\cite{CLIP} to incorporate visual aspects of text into font generation~\cite{Izumi2024CLIPFontDraw, FontCLIP}.
Currently, GRIF-DM, proposed by Kang \etal~\cite{GRIF-DM}, is the state-of-the-art impression-guided font generation method, where impression vectors derived from BERT~\cite{BERT} are used as conditions in a diffusion model.
In this paper, we develop a diffusion model based on GRIF-DM to demonstrate that the impression vectors derived from our method can be effectively used as conditions for high-quality font generation.
\par

\subsection{Word Embedding Models and LLMs}
Large language models (LLMs) have become ubiquitous for vectorizing text in various tasks, owing to their training on massive corpora.
Seminal examples include word2vec and BERT, which remain widely used for both standalone text processing~\cite{Al2020Word2vec, Min2023Recent} and multimodal applications such as image captioning~\cite{Leibem2016Visual, Wang2020Cross, Li2020Oscar} and text-based image generation~\cite{Rombach2022High, Ramesh2022Hierarchical, Frans2022CLIPDraw}. 
In the latter domain, CLIP~\cite{CLIP} has garnered particular attention by aligning images and textual captions in a shared latent space, making the learned text features widely applicable to vision-and-language tasks.
\par

Despite their versatility, these language models are not explicitly designed to handle the specialized scenario of font impressions. 
In practice, tags with distinct meanings in general usage can still apply to the same font. 
For instance, \cref{fig:font_ex} shows that \IT{happy} and \IT{funny} are assigned to the same font, indicating a strong relationship in terms of font impressions despite their differing general meanings. 
This paper compares our proposed method with BERT and CLIP, demonstrating that it captures font-related impressions more effectively than LLM-based embeddings.

\section{Font Impression Dataset}
\label{sec:dataset}
Throughout this paper, we use the MyFonts dataset~\cite{MyFonts}, collected from \href{https://www.myfonts.com/}{MyFonts.com}, one of the largest font marketplaces on the web.
\cref{fig:font_ex} shows several examples.
The original dataset contains 18,815 fonts and 1,824 impression tags, assigned by both font design experts and non-expert users in an open-vocabulary manner.
Consequently, the dataset inevitably contains missing and noisy tags.
\par

Since our goal is to analyze the impressions conveyed by font shapes, we exclude dingbat fonts (\ie, non-letter symbols) via manual inspection.
After this exclusion, 17,380 fonts remain, each associated with a subset of the 1,818 impression tags.
Each font has between 1 and 184 tags, with an average of 16.1 and a median of 11.
Each tag appears in 1 to 5,991 fonts, with an average of 153 and a median of 35.
In the following experiments, we split these 17,380 fonts into 15,639 for training and 1,741 for testing.
\par

\section{Embedding Impression Tags Based on Co-occurrence}
\subsection{Overview}
\cref{fig:overview} illustrates an overview of our proposed embedding method based on spectral embedding~\cite{spectral}.
Specifically, we first construct a graph based on the co-occurrence of impression tags across fonts, where each node corresponds to an impression tag.
An edge between two nodes is weighted according to their \emph{conditional co-occurrence probabilities} in the dataset. 
We then represent the graph as a weighted adjacency matrix and perform an eigenvalue decomposition on its Laplacian matrix to embed each node into a low-dimensional embedding space.
This low-dimensional representation is referred to as the spectral embedding, which we use as the vector representation for each impression tag.
\par

\subsection{Co-occurrence of Font Impression Tags}
Assume we have a collection of fonts, in which each font comprises a set of character images (\eg, `A'–`Z') in a particular style. 
As shown in \cref{fig:font_ex}, each font is assigned multiple impression tags.
In most cases, impression tags are single words (\eg, \IT{cool}), but some are hyphenated phrases (\eg, \IT{comic-book}). 
Let $N$ denote the vocabulary size of the impression tags.
\par

The co-occurrence of two impression tags means that both tags are assigned to the same font.
For instance, in the left part of \cref{fig:overview}, the tags $t_1$ and $t_2$ co-occur in font $F_1$, and the same pair also co-occurs in fonts $F_2$ and $F_4$. 
By contrast, $t_1$ and $t_3$ co-occur only in $F_1$. 
By counting how often each tag pair co-occurs across all fonts, we construct a graph.
The graph’s nodes correspond to the overall vocabulary of impression tags, and each edge is weighted according to their co-occurrence in training fonts.
\par

More specifically, each edge weight is computed as the average of the conditional co-occurrence probabilities between the two tags (nodes).
Given two tags $t_i$ and $t_j$, where $i, j \in [1, N]$, we calculate $P(i \mid j)$ and $P(j \mid i)$ by dividing the co-occurrence count by the individual frequencies of $t_j$ and $t_i$, respectively.
Finally, the edge weight between $t_i$ and $t_j$ is defined as $(P(i\mid j)+P(j\mid i)) / 2$.
\par

A simpler approach might be to use raw co-occurrence frequencies as edge weights.
However, impression tags vary widely in frequency; some are common (high-frequency), while others are rare (low-frequency).
Relying solely on raw frequency would make high-frequency tags appear strongly correlated with nearly all others, obscuring the nuanced relationships involving rare tags.
Instead, by averaging the conditional co-occurrence probabilities, we better capture specific associations for infrequent tags and reduce the bias toward dominant ones.
\par

\subsection{Graph Spectral Embedding of Impression Tags}
\label{sec:graph-spectrum}
As shown in \cref{fig:overview}, we use spectral embedding on the co-occurrence graph to represent each impression tag as a vector.
Specifically, to compute the spectral embedding vectors, we perform eigenvalue decomposition on the Laplacian matrix of the co-occurrence graph.
First, we construct an $N \times N$ adjacency matrix that defines the edge weights between nodes, and then compute the corresponding $N\times N$ symmetric normalized Laplacian matrix.
Next, we perform eigenvalue decomposition on this matrix and select the first $d$ eigenvectors corresponding to the $d$ smallest eigenvalues.  
\par

We represent each impression tag $t_i$ by stacking the $i$-th component of the first $d$ eigenvectors and normalizing the resulting vector.
More specifically, let $\mathbf{v}_1, \dots, \mathbf{v}_d$ be the top $d$ eigenvectors of the Laplacian matrix, each of length $N$ (number of tags).
We obtain the $d$-dimensional embedding for impression tag $t_i$ by extracting the $i$-th component from each eigenvector and stacking them into the vector $\bigl(\mathbf{v}_1(i), \dots, \mathbf{v}_d(i)\bigr)^\top$.
Finally, we normalize this vector to have unit length.
\par

Our method relies on a linear-algebra-based approach, in contrast to neural network-based embeddings such as BERT or CLIP.
One might assume that a linear approach has limited expressive power; however, our experiments show that this spectral embedding strategy outperforms embeddings obtained from BERT or CLIP for capturing font impressions.
Moreover, our method avoids dependence on complex neural architectures and iterative training procedures.
\par

Using the training set of $15,639$ fonts introduced in \cref{sec:dataset}, we embed $1,818$ impression tags into a $d$-dimensional space following the procedure described above. 
We set $d=100$ based on the analysis provided in \cref{sec:determining-d}. 
In the experiments in later sections, these $1,818$ embedding vectors (each of dimension 100) serve as the numerical representations of individual impression tags.
\par

\subsection{Determining the Embedding Dimension}
\label{sec:determining-d}

\begin{figure} 
    \centering
    
    \begin{subfigure}{0.52\linewidth}
        \includegraphics[width=\linewidth]{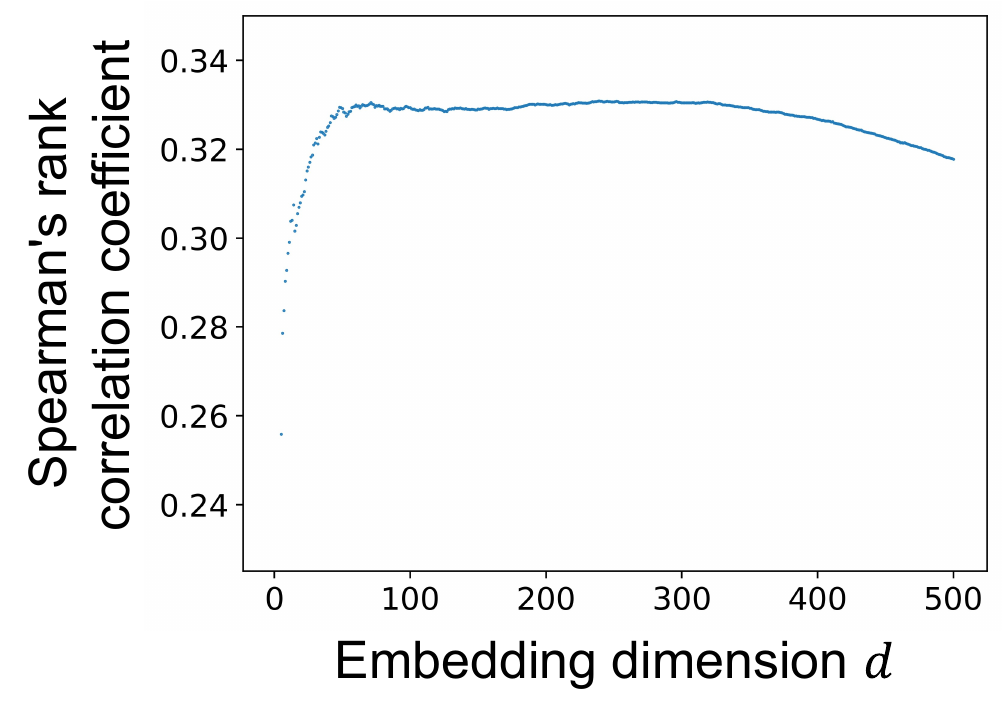}
        \caption{}
    \end{subfigure}
    \hfill
    \begin{subfigure}{0.47\linewidth}
        \includegraphics[width=\linewidth]{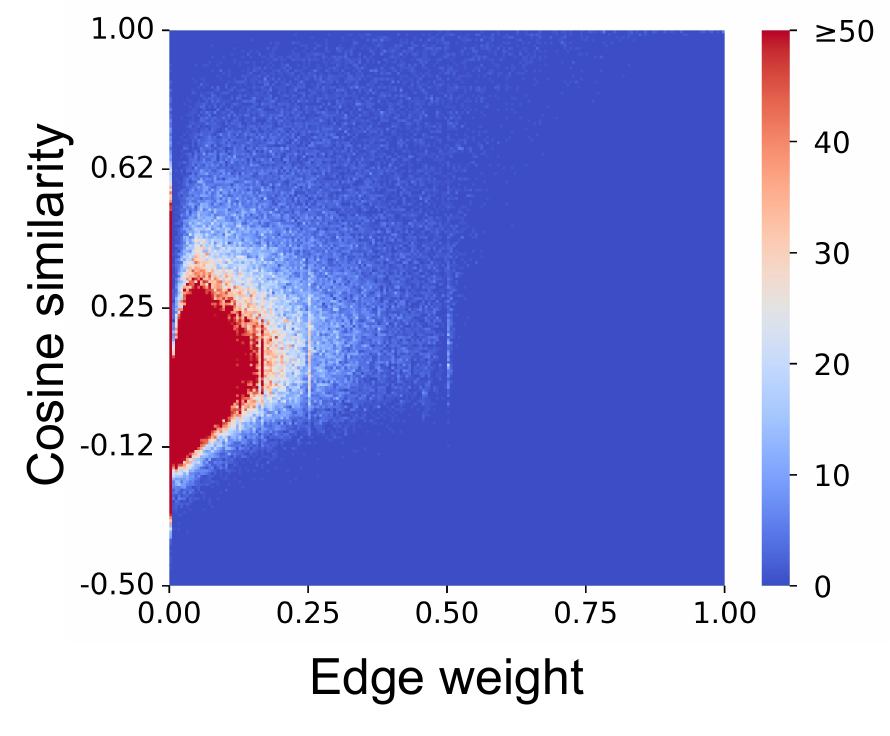}
        \caption{}
        \label{fig:short-b}
    \end{subfigure}
    \caption{We determine $d$ so that frequently co-occurring tags are placed closer together in the embedding space. (a)~Spearman’s rank correlation coefficient for each value of $d$. (b)~Correlation between edge weights and the cosine similarity of embedding vectors for tag pairs, with dimensionality $d=100$.}
    \label{fig:determing-d}
\end{figure} 

Choosing the number of eigenvectors $d$ (\ie, the embedding dimension) is critical.
Smaller values of $d$ lead to insufficient representational capacity, whereas very large values result in sparse representations that obscure relationships between tags.
We determine $d$ based on the principle that tags connected by larger edge weights should be placed closer together in the embedding space.
Specifically, we vary $d$, compute Spearman’s rank correlation coefficient between the edge weight of each tag pair and the cosine similarity of their embedding vectors, and then select the dimensionality that maximizes this correlation.
\par

\cref{fig:determing-d}(a) shows the correlation coefficients for each dimension $d$, computed between the edge weights and the cosine similarities of the embedding vectors.
The figure indicates that the correlation peaks in the range of 100 to 300 dimensions and changes only slightly within this range, suggesting that 100 dimensions are almost as effective at representing the data as 300.
Therefore, we set $d=100$ in the subsequent experiments.
Despite this relatively small dimension, our method outperforms BERT and CLIP, as demonstrated in the experimental results presented later.
\par

\cref{fig:determing-d}(b) provides a heatmap comparing each tag pair’s edge weight in the co-occurrence graph with the cosine similarity of their embedding vectors. 
The heatmap indicates a positive correlation: pairs of tags with larger edge weights generally exhibit higher cosine similarity, indicating that the embedding vectors effectively capture co-occurrence relationships.
Note that most edge weights are small, suggesting that only a limited subset of tag pairs strongly co-occur.
In contrast, some pairs with low edge weights still display relatively high similarity.
This situation arises when two tags do not directly co-occur but are indirectly related.
For example, tags A and B co-occur, and tags B and C also co-occur, then, even if tags A and C never co-occur (and therefore, their edge weight is zero), their embedding vectors become closer.
\par

\section{Distribution of Impression Vectors}
\label{sec:observation}
\subsection{Observing the Distribution by t-SNE}

\begin{figure} 
    \centering
    
    \begin{subfigure}{0.325\linewidth}
        \includegraphics[width=\linewidth]{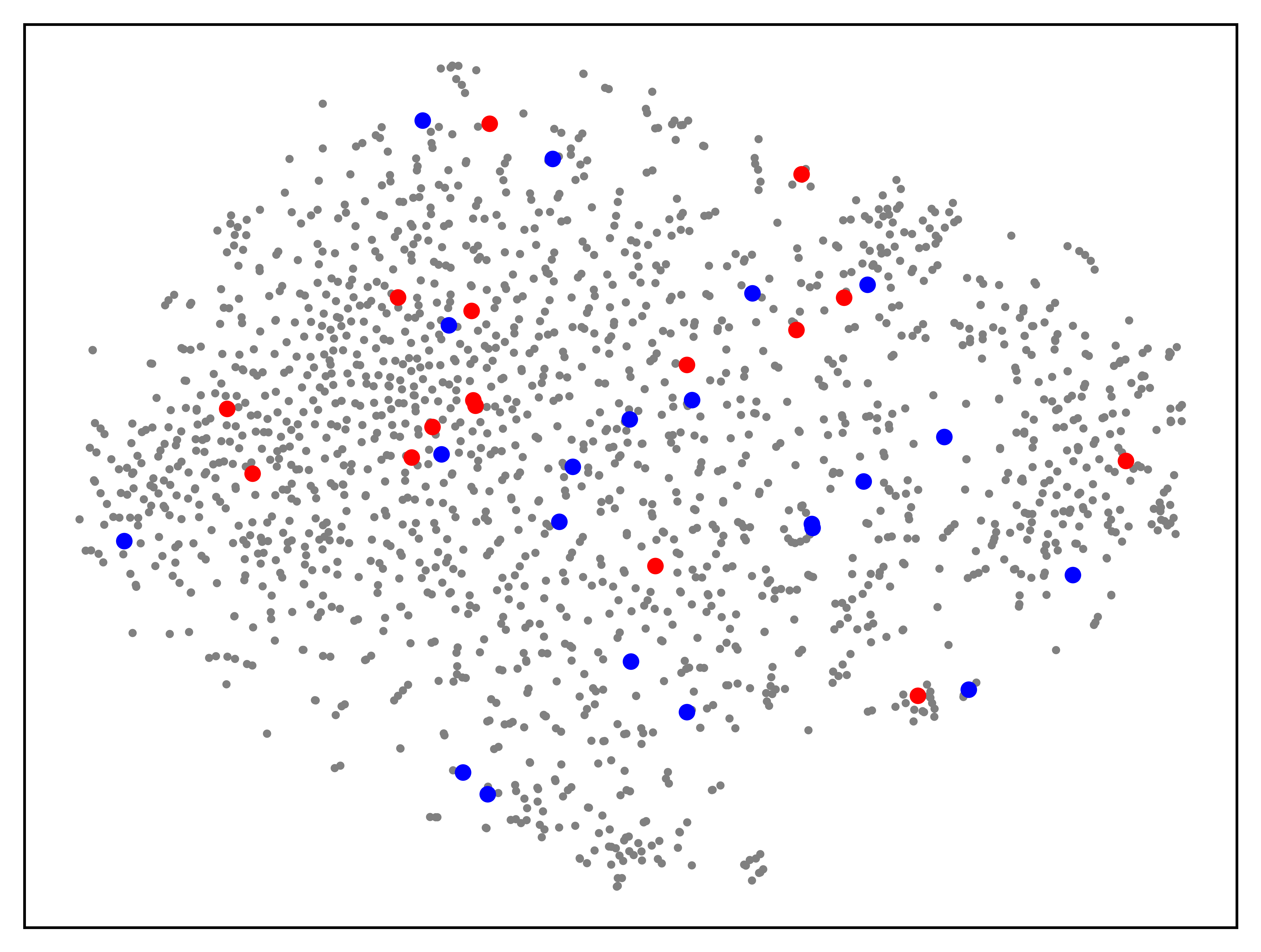}
        \caption{BERT}
        \label{fig:tSNE_BERT}
    \end{subfigure}
    \hfill
    \begin{subfigure}{0.325\linewidth}
        \includegraphics[width=\linewidth]{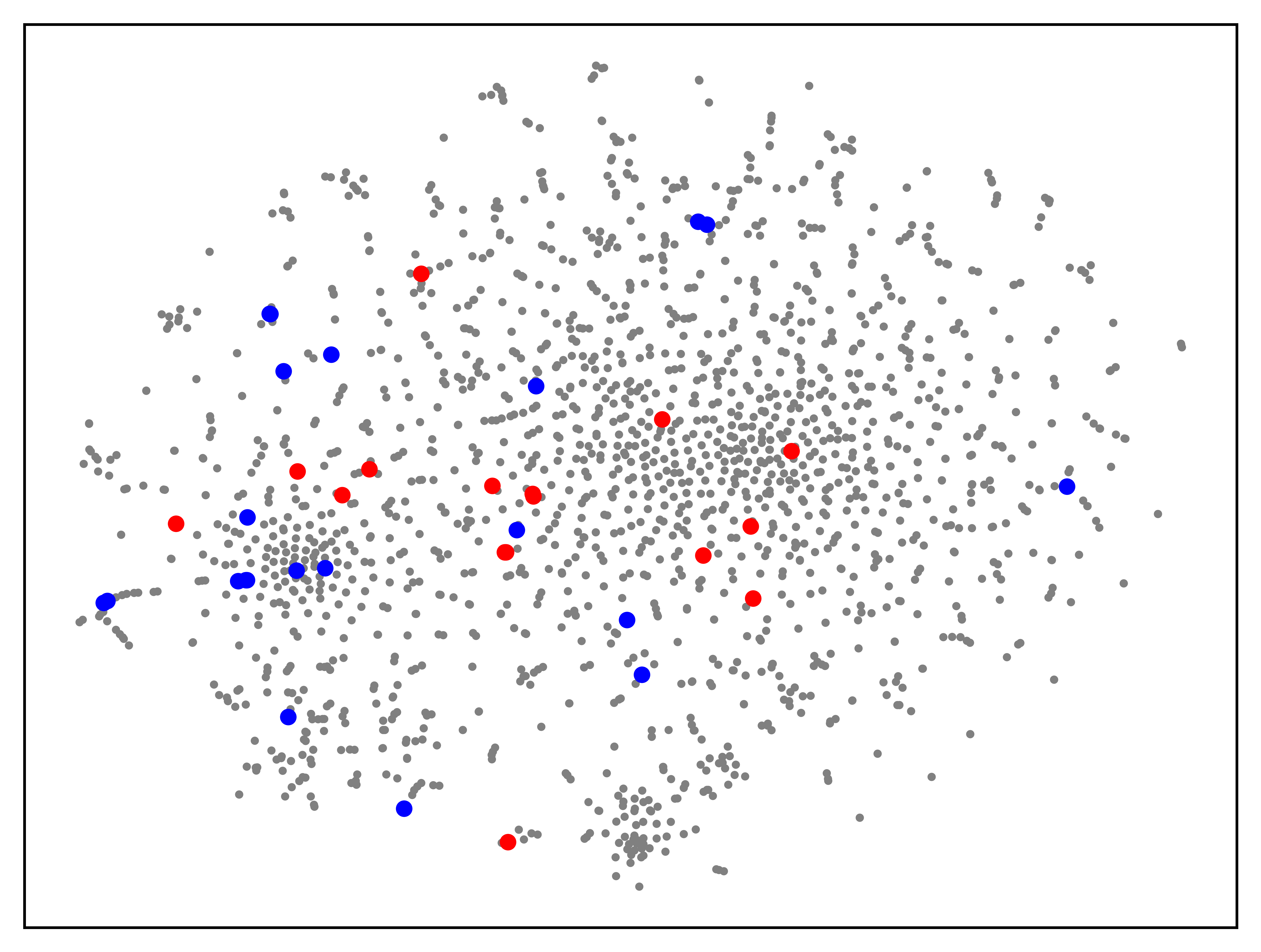}
        \caption{CLIP}
        \label{fig:tSNE_CLIP}
    \end{subfigure}
    \hfill
    \begin{subfigure}{0.325\linewidth}
        \includegraphics[width=\linewidth]{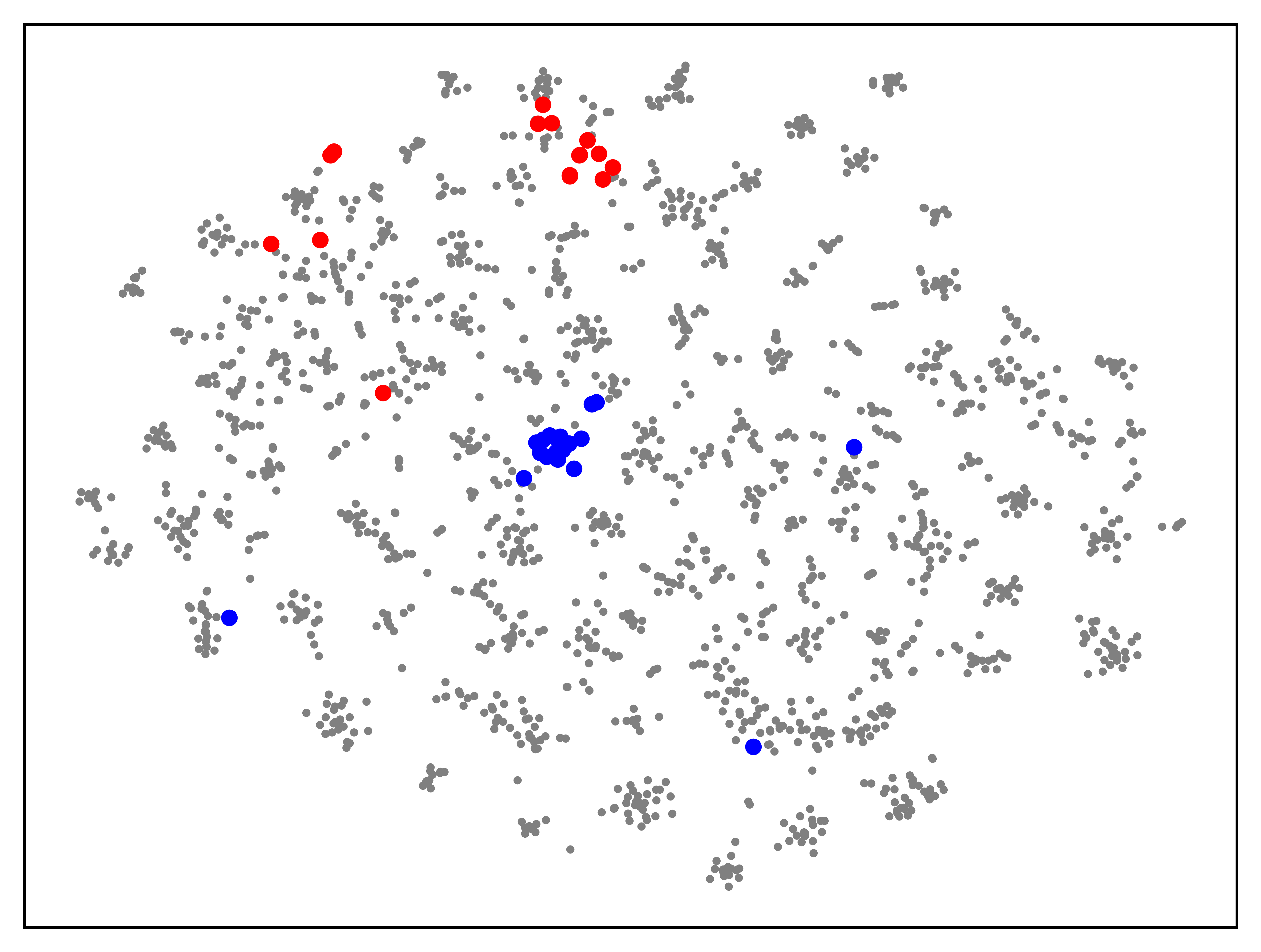}
        \caption{Proposed}
        \label{fig:tSNE_spectral}
    \end{subfigure}
    
    \caption{t-SNE visualization of the embedding vector distributions. The red and blue dots represent impression tags that belong to the same cluster respectively, in the proposed embedding space, as listed in \cref{fig:cluster_ex}.}
    \label{fig:tsne-distribution}
\end{figure} 

We compared the properties of our impression tag embeddings described in \cref{sec:graph-spectrum} with those derived from BERT~\cite{BERT} and CLIP~\cite{CLIP}. 
To obtain the embedding vector for each tag using BERT, we simply input each tag as a single-word prompt and extracted its embedding vector using the publicly available checkpoint \texttt{google-bert/bert-base-uncased}. 
Similarly, we extracted embedding vectors from CLIP using the checkpoint \texttt{openai/clip-vit-base-patch32}.
 \par

\cref{fig:tsne-distribution} visualizes the distributions of impression tag embedding vectors obtained using BERT, CLIP, and our proposed method. 
In the BERT and CLIP embedding spaces, the vectors appear to be distributed uniformly, whereas in the proposed method, the distribution exhibits non-uniform density. 
Notably, the proposed method forms numerous small clusters, which set it apart from the other two methods. 
This indicates that, by explicitly accounting for each tag’s relationship to font images, our approach produces embeddings that are substantially different from those of text-based methods.
Moreover, the formation of numerous small clusters suggests that certain impression tags frequently co-occur, while others rarely do. 
\par

\subsection{Clustering the Impression Vectors}
Since the t-SNE visualization revealed the cluster structure of the vectors produced by the proposed method, we applied standard $k$-means clustering to all $N=1,818$ impression tags in the $100$-dimensional embedding space. 
The number of clusters, $k$, was set to $140$ based on the peak of the silhouette score, a commonly used clustering quality metric.
\par

In \cref{fig:tsne-distribution}(c), the red and blue dots represent impression tags that belong to the same cluster in the proposed embedding space.
Although the red dots appear slightly scattered due to the nonlinearity of t-SNE, the blue dots form a clear cluster. 
An important observation is that the same tags appear more widely scattered in (a) and (b), indicating that the proposed embedding method exhibits different characteristics from those of BERT and CLIP. 
\par

\begin{figure}[t] 
    \centering
    \includegraphics[width=\linewidth]{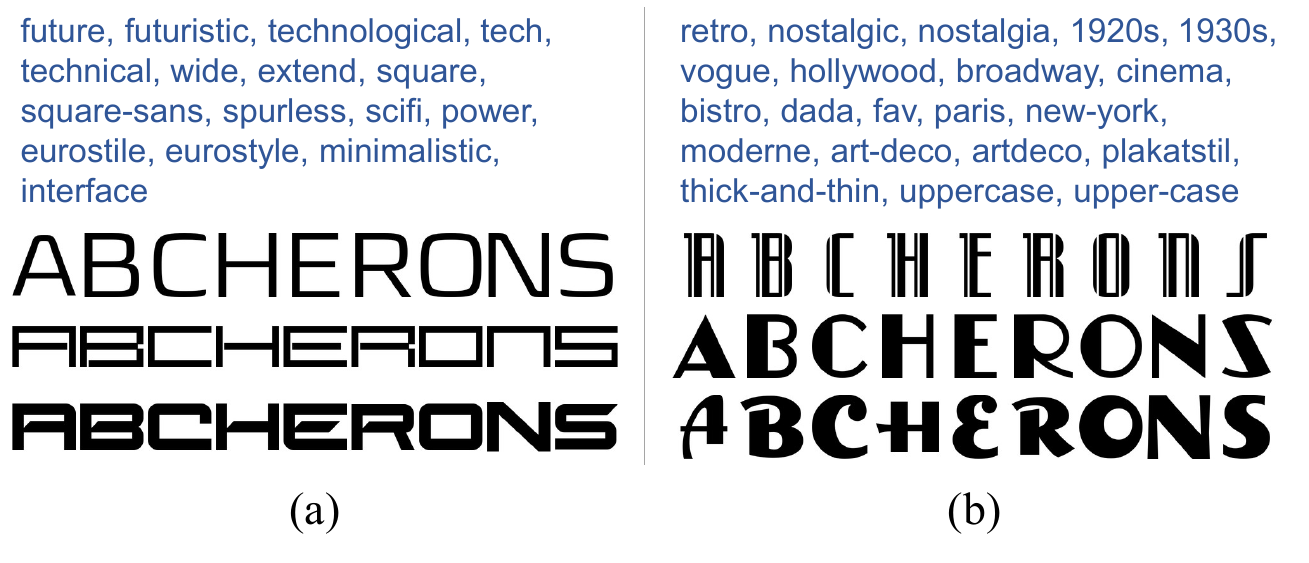}
    \vspace{-20pt}
    \caption{Examples of tag clusters. The tags in (a) correspond to the red dots in \cref{fig:tsne-distribution}, and those in (b) correspond to the blue dots. Shown below each cluster are three example fonts that are closely associated with the corresponding tags.}
    \label{fig:cluster_ex}
\end{figure} 

\subsection{Cluster-Wise Observation}
\figsref{fig:cluster_ex}(a) and (b) show two example tag clusters along with sample fonts that have multiple tags that belong to each cluster. 
First, looking at the tag clusters themselves, we find that even though the proposed method is based solely on tag co-occurrence, the resulting  embedding vectors of tags reflect semantic similarity in natural language and group similar tags into the same cluster (\eg, \IT{future} and \IT{futuristic} in (a) and \IT{nostalgic} and \IT{nostalgia} in (b)). 
This observation indicates that the proposed method works effectively. 
Furthermore, when examining the fonts associated with these clusters, we observe that the fonts in (a) exhibit fairly consistent stroke widths, whereas in (b), they are more varied. 
Hence, each cluster is underpinned by a set of fonts that share certain style features.
\par

On the other hand, there are cases where the clusters do not align with a tag’s typical meaning in natural language. 
For instance, in tag cluster~(a), the tags \IT{technological}, \IT{wide}, \IT{extend}, and \IT{square} are grouped with \IT{future} although the word \IT{future} has a significantly different meaning from the others in natural language. 
However, when we consider font shapes, these tags are reasonably similar. 
Indeed, fonts labeled with \IT{future} often have wide, square shapes. 
It is also notable that the abstract word \IT{future} is associated with the geometric tags \IT{wide} and \IT{square}. 
Similarly, in cluster~(b), \IT{thick-and-thin} is grouped with \IT{retro} and \IT{nostalgic}, suggesting that a classic impression is often conveyed by fonts with contrasting thick and thin strokes. 
These results demonstrate that the proposed method can derive impression vectors that accurately reflect actual font shapes, despite relying solely on co-occurrence.
\par

\section{Relationship between Embedded Impression Tags and Font Images}
\label{sec:relationship}

\subsection{Consistency between Impressions and Fonts}
\label{sec:consistency}
Fonts with similar shapes should share similar impressions; conversely, similar impression tags should be assigned to similarly shaped fonts. We refer to this property as the \emph{consistency} between impression tags and fonts. 
Our tag-embedding method needs to reflect the consistency; namely, impression tags that are close to each other in the embedded space should be assigned to the same fonts.
Specifically, consider an impression tag $t_i$ (where $i = 1, \dots, 1,818$) and let $F_{t_i}$ be the set of test fonts associated with this tag. 
We then find the tag $s_{i}$ closest to $t_i$ in the embedding space and obtain the corresponding font set $F_{s_{i}}$. 
If $t_i$ is indeed similar to $s_{i}$, then the set $F_{t_i}$ should also be similar to $F_{s_{i}}$.
\par

\Cref{tab:quantitative} quantifies this consistency by measuring the similarity between the two font sets, $F_{t_i}$ and $F_{s_{i}}$.  
We used three set similarity metrics: the Jaccard coefficient, the Dice coefficient, and the overlap coefficient.
The Jaccard coefficient represents the ratio of the intersection to the union of two sets, the Dice coefficient emphasizes common elements, and the overlap coefficient measures the ratio of the intersection to the size of the smaller set. 
All three metrics yield values in the range $[0, 1]$, where a higher value indicates greater similarity. 
The values shown in the table are averaged over all $t_i$.
\par

Our proposed method outperforms BERT and CLIP across all metrics, indicating that font sets associated with closely embedded impression tags tend to overlap more. 
In particular, this difference is especially pronounced for the overlap coefficient. 
Since the overlap coefficient uses the smaller of the two sets as its reference for measuring the intersection, it is more sensitive to low-frequency tags (those associated with fewer fonts). 
This dataset contains many low-frequency tags, mainly due to the open-vocabulary tag assignment process and missing tags. 
Nevertheless, when embedded using our method, even these low-frequency tags show more substantial overlap with the font sets of nearby tags.
This result suggests that our co-occurrence-based approach captures shape-based relationships among impression tags more accurately than text-based BERT and the general-purpose image-text alignment of CLIP.
\par

\subsection{Robustness to Missing and Noisy Tags}
\label{sec:robustness}
An appropriately constructed embedding of impression tags can help mitigate missing and noisy tags.
In the embedding space, tags positioned close to a given tag $t_i$ can be regarded as candidate tags that could reasonably be applied to the same font as $t_i$.
By assuming that all relevant neighbors should also be assigned to the font, we can reduce the risk of missing tags.
Conversely, if a font includes a tag that is embedded far away from its other tags, this outlier tag can be considered a noisy tag that may have been inappropriately assigned.
\par

\begin{table}[t] 
    \centering
    \caption{Quantitative evaluation of the consistency between the impression embedding space and the font image distribution, measured by the Jaccard, Dice, and Overlap coefficients. Robustness to missing and noisy tags is also evaluated.}
    {
        \fontsize{9pt}{11pt}\selectfont
        \begin{tabular}{
        p{1.28cm}  
        >{\centering\arraybackslash}p{0.85cm} 
        >{\centering\arraybackslash}p{0.7cm} 
        >{\centering\arraybackslash}p{1.1cm}
        | 
        >{\centering\arraybackslash}p{0.95cm} 
        >{\centering\arraybackslash}p{0.85cm}}
            \toprule
            Method &  Jaccard$\uparrow$ & Dice$\uparrow$ & Overlap$\uparrow$ & Missing$\downarrow$ & Noisy$\uparrow$\\
            \midrule
            BERT \cite{BERT}        & 0.023 & 0.037 & 0.091 & 790.2 & 0.122 \\
            CLIP \cite{CLIP}        & 0.044 & 0.068 & 0.175 & 745.3 & 0.163 \\
            Proposed       & \textbf{0.139} & \textbf{0.198} & \textbf{0.344} & \textbf{543.5} & \textbf{0.594}\\
            \bottomrule
        \end{tabular}
    }
    \label{tab:quantitative}
\end{table} 

We evaluated whether the proposed method produces an embedding space capable of addressing missing and noisy tag problems. 
First, we assessed the method’s robustness to missing tags based on the proximity of tags assigned to each font in the embedding space.
Suppose a font is associated with $I$ distinct impression tags.
For each ordered pair $(t_i, t_j)$ of these tags (where $i, j \in [1, I]$, $i \neq j$), we determine the rank $p \in [1, N]$ of $t_j$ among $N-1$ impression tags based on its similarity to $t_i$.
We then average the rank $p$ over all $I(I - 1)$ ordered pairs.
\par

We define this average value as the font's ``missing-tag robustness,'' where smaller values indicate greater robustness. 
In the ideal---but practically unachievable---scenario, every tag pair assigned to a font would be each other’s nearest neighbor, resulting in an average of $1$. 
We then compute the mean of this value across all test fonts to evaluate  the embedding method’s missing-tag robustness.
\par

Next, we quantify robustness to noisy tags by assuming that a tag that is relatively dissimilar to the others can be considered a noisy tag.
Given a set consisting of $I$ tags assigned to a font and one additional tag randomly selected from the $N$ total tags, we compute the average distance between each of the $I+1$ tags and the remaining $I$ tags.
If the randomly selected tag yields the largest average distance, we consider it successfully identified as a noisy tag.
We repeat this process for each font using ten randomly selected additional tags for each test font, and compute the proportion of trials in which the random tag is identified as a noisy tag.
We define this proportion as the embedding method’s noisy-tag robustness, with higher values indicating stronger robustness. 
In the ideal case---which, again, is unlikely in practice---this proportion would be $1$, meaning that the random tag is always detected as a noisy tag in every trial.
\par

\Cref{tab:quantitative} presents the robustness of each embedding space to missing and noisy tags. 
An embedding space is considered robust if it enables the detection of missing tags (overlooked) and noisy tags (erroneous). 
The results show that the proposed method achieves greater robustness to both missing and noisy tags.
Hence, it is more effective in handling incomplete or erroneous font-impression annotations. 
This suggests that our co-occurrence-based approach, which grounds impression tags in font shapes, yields a clearer representation of semantic proximity, thereby facilitating more accurate tag correction and supplementation.
\par

\section{Font Generation Using Impression Tags}
\label{sec:font_gen}

\subsection{Impression-Guided Font Generation with a Diffusion Model}
\begin{figure}[t] 
    \centering
    \includegraphics[width=\linewidth]{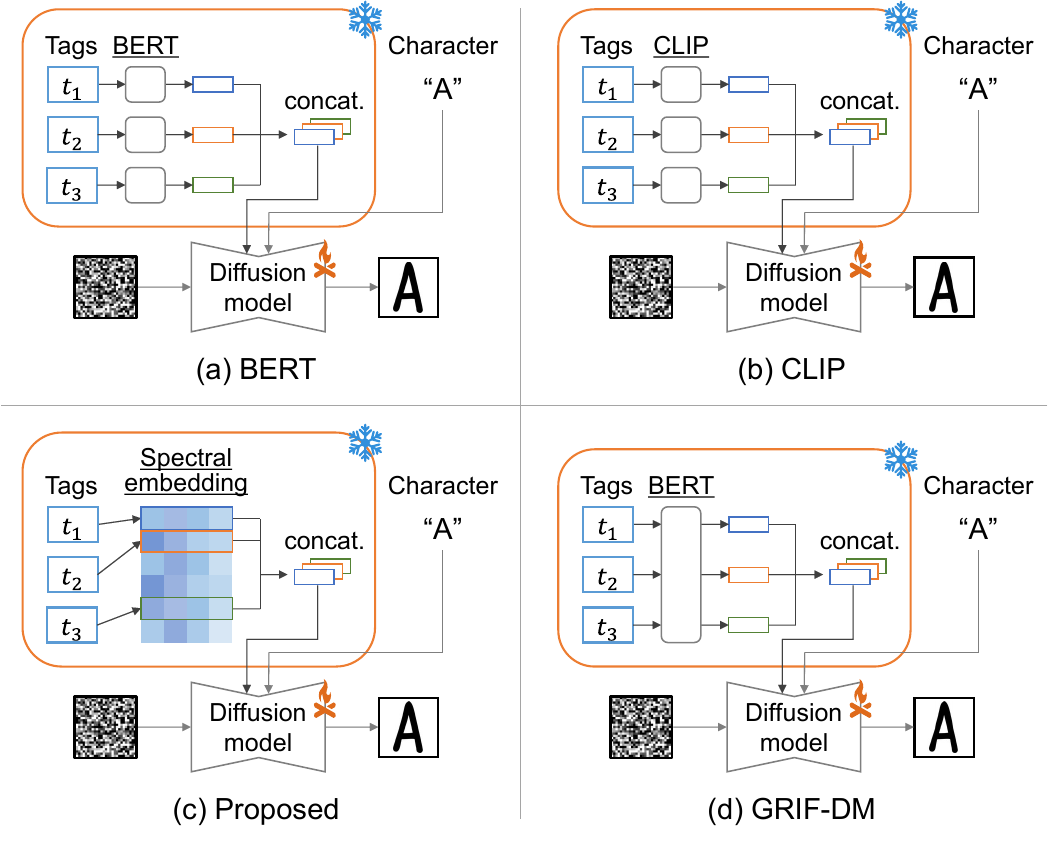}
    \caption{Impression-guided font generation using different word embedding methods.}
    \label{fig:diffusion_overview}
\end{figure}

Once impression tags are embedded as vectors, one natural application is to generate fonts that reflect a given impression. 
As described in \cref{sec:review-gen}, several methods have already pursued this goal, with GRIF-DM~\cite{GRIF-DM}, which represents the state-of-the-art diffusion-based approach.
\cref{fig:diffusion_overview}(d) illustrates the architecture of GRIF-DM, in which a set of impression tags is first vectorized by BERT and then used to condition the diffusion model.
\par

Building on the concept of GRIF-DM, we replace its impression-conditioning vectors with those derived from BERT, CLIP, or our proposed method, as illustrated in \figsref{fig:diffusion_overview}(a), (b), and (c), respectively. 
It is worth noting that in both (a) and (d), BERT serves as the underlying language model, but the impression vectors are obtained in different ways: in (a), each tag is individually embedded, whereas in (d), all tags are input collectively as a single prompt.
We expect the latter approach to capture tag interrelationships more effectively through  BERT’s self-attention mechanism.
\par

We also compared our impression embedding method with the state-of-the-art approach, Impression-CLIP~\cite{ImpressionCLIP}.
This method associates fonts with impressions via CLIP-based contrastive learning.
Impression-CLIP embeds fonts and impressions into a shared latent space such that positive font–impression pairs are closer and negative pairs are farther apart.
We trained Impression-CLIP using the same training fonts and tags as our proposed method. 
We then used its impression encoder to vectorize impression tags and condition the diffusion model.
We compared its font generation performance with that of our proposed method.
\par

\subsection{Experimental Setup}
To train each diffusion model, we used the 15,639 training fonts introduced in \cref{sec:dataset}. 
Specifically, for each training font, the model was trained to generate images of the 26 uppercase letters ‘A’–‘Z’ by conditioning on the font’s impression-tag set and a target letter. 
Other training details follow those of the original GRIF-DM paper~\cite{GRIF-DM}.
\par
We evaluated the quality of generated images using the Fr\'echet Inception Distance (FID) and the Structural Similarity Index (SSIM). 
To investigate the effect of the number of impression tags $K$, we evaluated $K \in \{1,3,5,7,10\}$ and the case $K=\mathrm{all}$ (\ie, all impression tags for a font).
Consequently, we focus on fonts that have at least ten impression tags and randomly select 5{,}000 such fonts from the full set of 17{,}380 fonts for evaluation.
\footnote{%
As in GRIF-DM~\cite{GRIF-DM}, we combine training and test sets for FID evaluation to ensure a sufficient number of fonts.
Note that the trends in the evaluation results remain consistent even when using only the test set.
} 
For each of 5,000 fonts, we randomly shuffle the complete set of its impression tags. 
Then, for $K = 1$, we use the first tag from the shuffled list as a condition; for $K = 3$, the first three tags; and so on.
Thus, when $K=5$, we simply use the first five tags from the shuffled list.
\par

Regardless of $K$, the 5{,}000$\times$26 font images in this evaluation serve as ground-truth references. 
We compute the FID between the distribution of the 5{,}000$\times$26 ground-truth images and that of the 5{,}000$\times$26 generated images conditioned on $K$ impression tags per font. 
For SSIM, we compare each generated image directly with its corresponding ground-truth image and average the similarity scores across all 26 letters and all fonts.
\par

\subsection{Experimental Results}
\begin{figure} 
    \centering
    
    \begin{subfigure}{0.49\linewidth}
        \includegraphics[width=\linewidth]{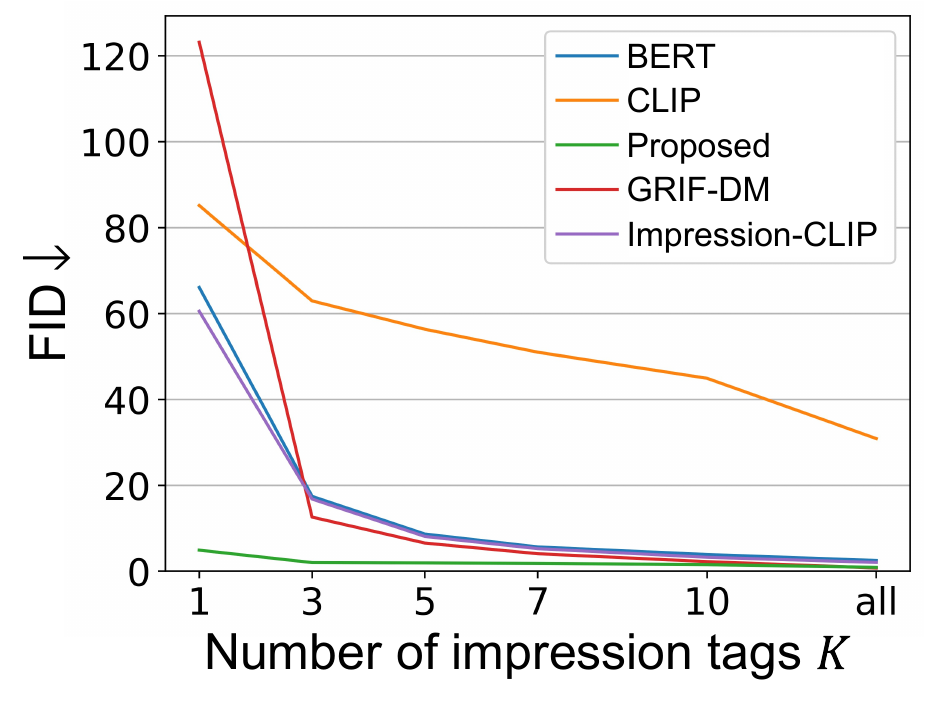}
    \end{subfigure}
    \hfill
    \begin{subfigure}{0.49\linewidth}
        \includegraphics[width=\linewidth]{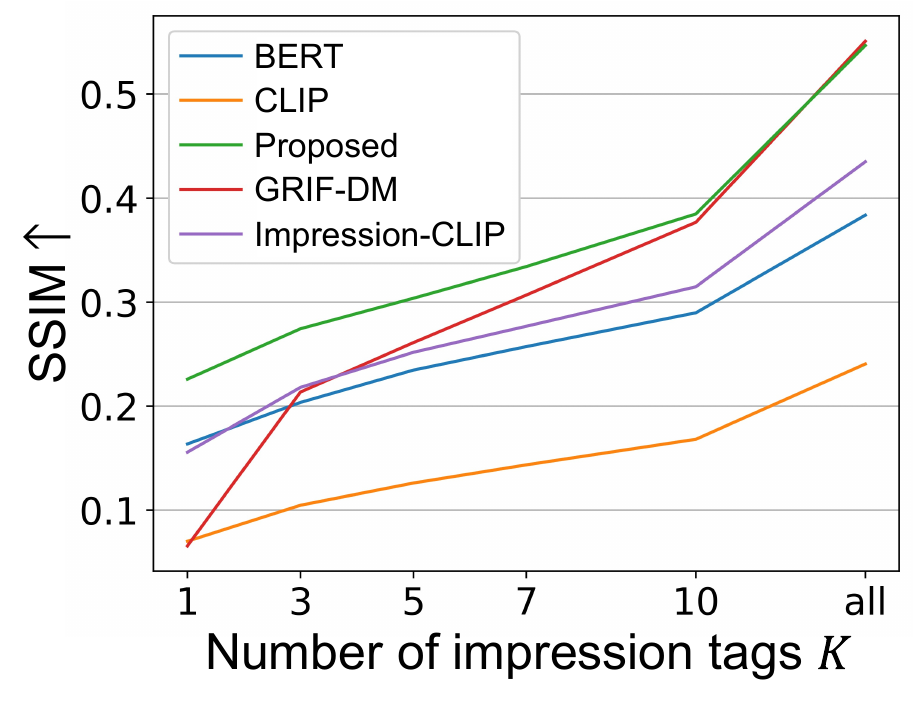}
    \end{subfigure}

    \caption{Quantitative evaluation of the quality of impression-guided font generation for different numbers of impression tags, $K$. ``all'' on the horizontal axis indicates that all impression tags associated with a font were used for generation.}
    \label{fig:NumTag_quant}
\end{figure}

\cref{fig:NumTag_quant} shows a quantitative evaluation of the generated font images across different values of $K$. 
Overall, the results indicate that using more impression tags generally improves the quality of the generated fonts. 
This is because supplying additional tags provides a more fine-grained description of the font’s shape, thus facilitating the generation of fonts closer to the ground truth. 
When comparing the proposed method with BERT and CLIP, the proposed method consistently outperforms both across all values of $K$. 
As discussed in \cref{sec:consistency}, this superiority can be attributed to the tighter alignment of the proposed impression-tag embedding space with the font-image distributions, which facilitates impression-based font generation.  
In other words, the co-occurrence-based vectors appear more effective than the general-purpose embeddings from BERT and CLIP.
Furthermore, our proposed method also outperforms the comparative embedding method, \ie, Impression-CLIP~\cite{ImpressionCLIP}.
\par

When focusing on GRIF-DM, the proposed method performs comparably to GRIF-DM when all tags are used, showing only a slight performance difference.
However, when fewer tags are available ($K < 10$), the proposed method significantly outperforms GRIF-DM.
A possible explanation is that GRIF-DM heavily relies on a self-attention mechanism to capture relationships among multiple tags, which makes it more difficult to accurately represent the font shape when only a few tags are available. 
Indeed, with very few tags (particularly when $K=1$), we find that even the simpler BERT-based method---without any self-attention among tags---can outperform GRIF-DM.
\par

\begin{figure}[t] 
    \centering
    \includegraphics[width=\linewidth]{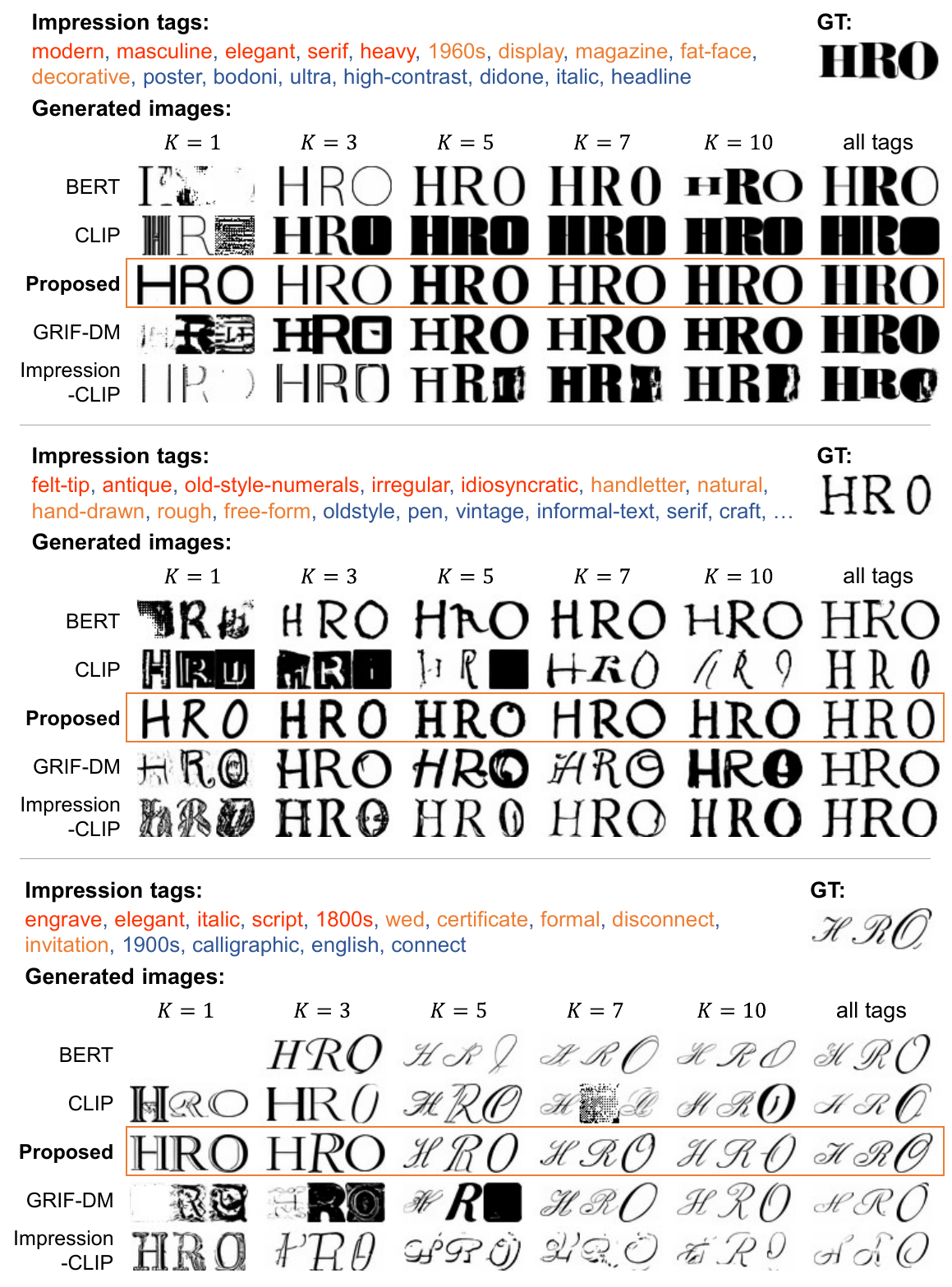}
    \caption{Generated images conditioned on different numbers of impression tags, $K$. In the top example, $K = 3$ indicates the use of the first three impression tags---\IT{modern}, \IT{masculine}, and \IT{elegant}---taken from those associated with the ground truth (GT). Other values of $K$ correspond to different numbers of tags. The first five and ten impression tags are highlighted in red and orange, respectively. All examples are taken from the test fonts.}
    \label{fig:gen_ex}
\end{figure}

\cref{fig:gen_ex} provides representative examples of the generated fonts. 
First, when all tags are used, both the proposed method and GRIF-DM generate fonts that closely resemble the ground truth across all samples. 
Next, when fewer tags are available, the other four methods---except for our proposed method---often fail to generate legible characters, whereas the proposed method, although its style deviates from the ground truth, still generates legible characters that reasonably match the available impression tags. 
In most examples, roughly $K=5$ tags are sufficient to produce a font style that closely resembles the ground truth, and using more tags further narrows the gap. 
For instance, in the top example with $K=5$, adding \IT{serif} and \IT{heavy} to the set \{\IT{modern}, \IT{masculine}, \IT{elegant}\} (\ie, $K=3$) makes the generated style closer to the ground truth. 
In real-world applications, it is not always possible to obtain a large set of impression tags.
Therefore, the ability to generate high-quality fonts from only a few tags is an important  requirement. 
Overall, the results demonstrate that the proposed method is effective even in practical scenarios where only a limited number of tags are available.
\par

\section{Conclusion}
This paper proposes a co-occurrence-based embedding method for font impression word tags, demonstrating that these embedding vectors are grounded in font shapes and more effectively capture the relationships between font shapes and impressions than general-purpose, text-centric models such as BERT and CLIP. 
Our experiments demonstrated that the proposed method is not only relatively robust to missing and noisy tags but also effective for impression-guided font generation. 
This practical value was confirmed in scenarios with both abundant and sparse impression tags.
\par

Despite these advantages, our method is limited to the existing impression-tag vocabulary, unlike LLMs that can embed arbitrary words. 
A promising direction is to integrate LLMs with our co-occurrence-based approach to vectorize arbitrary words consistently with font shapes.
Moreover, while our results suggest increased robustness to noisy or missing tags, these issues still pose challenges when co-occurrence data are severely incomplete.
Incorporating dedicated mechanisms to mitigate such labeling errors---similar to those used in prior work on handling noise~\cite{matsuda2022font}---represents another important avenue for future research.

\par
\bigskip
\noindent{\bf Acknowledgments}:\ This work was supported by JSPS KAKENHI-JP25H01149, JST CRONOS-JPMJCS24K4, JST BOOST-JPMJBS2406, and Hoso Bunka Foundation.

{
    \small
    \bibliographystyle{ieeenat_fullname}
    \bibliography{main}
}

\end{document}